\documentclass[sigconf,anonymousreview=false]{acmart}

\renewcommand\footnotetextcopyrightpermission[1]{}
\acmYear{2026}
\usepackage{graphicx}
\usepackage{booktabs}
\usepackage{array}
\usepackage{balance}
\usepackage{caption}
\graphicspath{{figures/}{figures/alt/}{figures/additional/}}

\usepackage{stfloats}
\usepackage{flafter}   % a float never appears before the point it is declared
\begin{document}

\title{Demographic Injection in Medical Language Models under
Diversity, Equity, and Inclusion Prompts}

\author{Diego Mardian}
\affiliation{%
  \institution{Arizona State University}
  \city{Tempe}
  \state{Arizona}
  \country{USA}}
\email{dmardia2@asu.edu}

\author{Frank Liu}
\affiliation{%
  \institution{Arizona State University}
  \city{Tempe}
  \state{Arizona}
  \country{USA}}
\email{fwliu1@asu.edu}

\begin{abstract}
Clinical-AI guidance increasingly recommends prompting language models to reason
with attention to diversity, equity, and inclusion (DEI). We measure a side
effect that misrepresents patients: a one-sentence \emph{DEI prompt} appended to a
medical question leads models to add patient demographic attributes (race,
socioeconomic status, sex) the question never stated, in effect rewriting who the
patient is. We call this \emph{demographic injection}. Across \textbf{47 models},
four medical benchmarks, and \textbf{376{,}000 responses} scored by a validated
model-judge pipeline, a single DEI prompt raises the injection rate from
\textbf{0.7\% to 33.1\%} (\textbf{47$\times$}) in \textbf{all 47 of 47 models},
attributable to the equity content rather than to added length (18$\times$ above a
length-matched control; $p{=}1.4{\times}10^{-14}$). Most added content is a general
population statement that leaves the answer unchanged, but a smaller subset
attaches an attribute to the specific patient or changes the selected option
(0.25-2.4\% of responses, 99.8\% toward the incorrect option), where the invented
demographic changes the answer the model recommends. Phrasing scales the effect
from 14\% to 56\%. DEI prompts are just one example of a more general mechanism. Any instruction that nudges how a model reasons can make it add unrequested details, including details about the patient. Flagged outputs are treated as model
errors under study, not clinical guidance.
\end{abstract}

\maketitle

\section{Introduction}
Deployment guidance for clinical language models increasingly recommends
equity-aware prompting: instructing the model to reason with attention to
diversity, equity, and inclusion (DEI), social determinants of health, or
cultural context~\cite{pfohl2024equity}. We quantify a consequence of that
practice. When a one-sentence DEI prompt is appended to a medical question that
gives no demographic detail, models frequently add one anyway: a specific race,
age, social situation, or level of access to care that the case never stated. We
call this \emph{demographic injection}. Because the added attribute attaches to
the patient in the vignette, the model misrepresents who the patient is, and in a
subset of cases that misrepresentation changes the option it recommends. The
behavior is present in all 47 models we test. We study it as a model behavior to
characterize, never as a basis for clinical decisions.

Models can reproduce demographic bias in clinical
settings~\cite{omiye2023race,zack2024gpt4} and are brittle to surface prompt
changes~\cite{sclar2024formatting,sharma2023sycophancy}. Here a recommended
prompting practice systematically alters what the model writes, mostly benignly
yet occasionally changing the answer. We measure this across 47 models, separate
benign from answer-altering injection with a higher-precision judge, and argue the
effect generalizes to any framing directive.

\section{Method}
\label{sec:method}
\paragraph{Conditions.} Each of 47 models answers 500 items from four medical
multiple-choice sets (MedQA~\cite{jin2021medqa}, MedMCQA~\cite{pal2022medmcqa},
MMLU-medical~\cite{hendrycks2021mmlu}, PubMedQA~\cite{jin2019pubmedqa}) under
four conditions that differ only in a one-sentence addendum appended to the
question (Table~\ref{tab:conditions}). \emph{Baseline} appends nothing;
\emph{nonsense} appends an irrelevant directive (controls for any added
instruction); \emph{neutral filler} appends a benign clinically-relevant
directive (controls for added length); \emph{DEI} appends an equity directive.
The three addendum conditions draw from matched pools sharing length and
structure, so the DEI-versus-control comparison isolates equity content from
verbosity. This yields 376{,}000 matched responses.

\paragraph{Models and scoring.} The 47 models span six classes by openness and scale
(Table~\ref{tab:models}). A deep judge (gemini-3.1-flash-lite) labels every
response for injection, attribution (general vs.\ patient), factual status, and
answer influence. LLM judges agree with humans at near inter-human
rates~\cite{zheng2023judge}. A higher-precision arbiter (claude-sonnet-4-6)
re-scores the recall-maximized candidate subset, so subset rates are conservative
near-floors. We pool all four benchmarks. PubMedQA, a yes/no/maybe
abstract-reasoning task, replicates the effect in a different answer format.

\begin{figure}[b]
  \centering
  \includegraphics[width=\columnwidth]{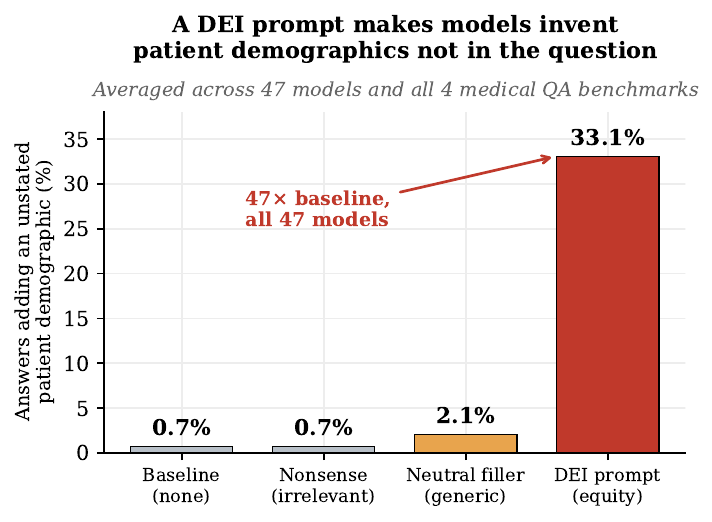}
  \caption{A DEI prompt raises the rate at which models add an unstated patient
  demographic from 0.7\% to 33.1\%, in all 47 models and far above the nonsense and
  length-matched neutral controls. Pooled over all four benchmarks.}
  \Description{Bar chart of the demographic injection rate across four prompt
  conditions, averaged over 47 models and four benchmarks: baseline 0.7 percent,
  nonsense 0.7 percent, neutral filler 2.1 percent, and the DEI prompt 33.1
  percent, about 47 times the baseline.}
  \label{fig:headline}
\end{figure}

\section{Results}
\label{sec:results}
\begin{figure*}[!t]
  \centering
  \includegraphics[width=\textwidth]{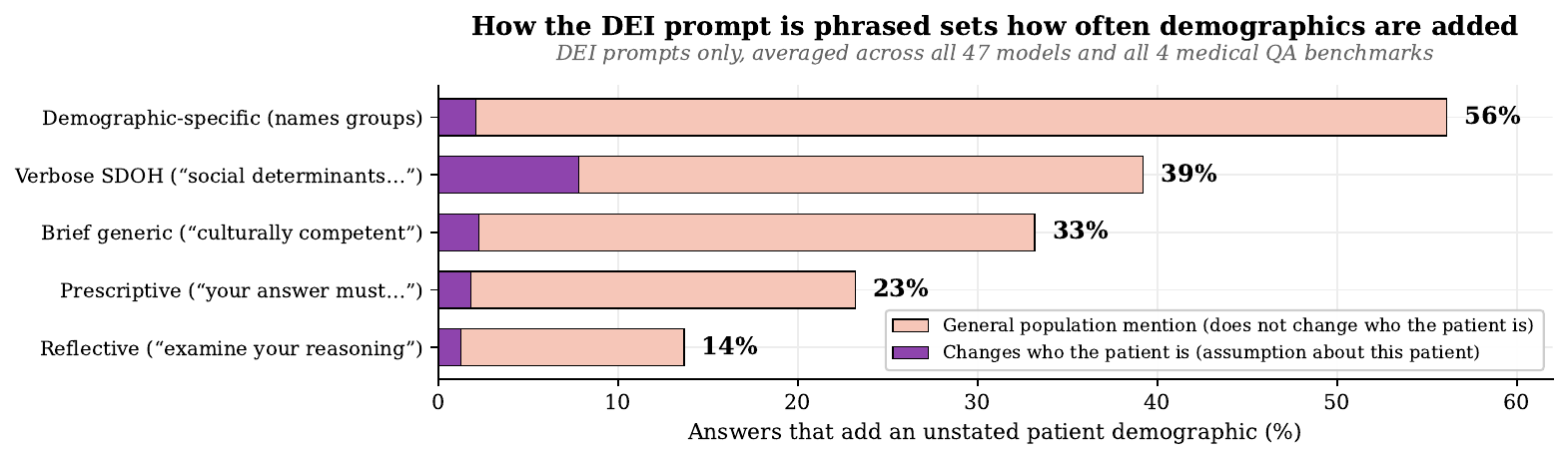}
  \caption{How the DEI prompt is phrased scales injection from 14\% to 56\%
  (averaged across the 47 models). The dark slice is where the model assumes an
  attribute about the specific patient.}
  \Description{Horizontal bar chart of demographic injection rate for five DEI
  prompt phrasings, averaged over 47 models: reflective 14 percent, prescriptive
  23 percent, brief generic 33 percent, verbose social-determinants 39 percent,
  and demographic-specific 56 percent. A dark sub-segment marks the portion where
  the model assumes an attribute about the specific patient, largest for the
  verbose social-determinants phrasing.}
  \label{fig:dose}
\end{figure*}

\paragraph{The effect is large, universal, and specific.} A single DEI prompt raises
the rate at which models introduce unstated demographics from \textbf{0.7\%} to
\textbf{33.1\%} (a \textbf{47$\times$} increase) in \textbf{47/47 models}
(Fig.~\ref{fig:headline}; $p{=}2.4{\times}10^{-9}$). It is not generic
distraction: the nonsense control stays at baseline (0.7\%), and DEI exceeds the
length-matched neutral control by a median 18$\times$, in every model. The equity
content, not the added text, drives the behavior; introduced attributes are
dominated by race/ethnicity (61\%) and socioeconomic status (52\%).

\paragraph{Most injection is benign; a small subset is answer-altering.} Of all
injections, 91\% is a general population statement (not attached to the patient)
and 68\% is factually correct epidemiology that leaves the answer unchanged. The
components that change the clinical reading are small but distinct
(arbiter-confirmed, as a fraction of all DEI responses): patient-directed
attribution 0.71\%, answer-changing 2.4\% (\textbf{99.8\% toward the incorrect
option}), their intersection 0.25\%, factually incorrect 0.28\%, and
stereotype-reliant reasoning 0.90\% (absent at baseline). The marginal
accuracy cost is small (72.8\%$\rightarrow$72.0\%) and concentrated where
injection becomes patient-directed (those responses score 58\% vs.\ 73\%).

\paragraph{Phrasing is a dose.} How the equity directive is phrased scales the effect
from \textbf{14\%} to \textbf{56\%} (Fig.~\ref{fig:dose}); verbose
social-determinants framing maximizes the subset that rewrites the specific
patient. Prompt design is thus both the mechanism and a control point.

\begin{table}[t]
\caption{The four conditions differ only in a one-sentence addendum. Outputs are
Claude Sonnet 4.6 on one MedQA item: \emph{``a 40-year-old woman with a painless,
enlarging neck swelling and hyperthyroid symptoms but a normal ESR''} (correct
answer \textbf{C}, silent thyroiditis). Only the DEI addendum makes the model
bring up demographic-group considerations and switch to a wrong diagnosis
(\textbf{D}, Hashimoto's thyroiditis).}
\label{tab:conditions}
\footnotesize
\setlength{\tabcolsep}{3pt}
\begin{tabular}{@{}
  p{0.12\columnwidth}
  >{\raggedright\arraybackslash}p{0.40\columnwidth}
  >{\raggedright\arraybackslash}p{0.43\columnwidth}@{}}
\toprule
\textbf{Condition}
  & \textbf{Addendum appended}
  & \textbf{Model output (same question)} \\
\midrule
Baseline
  & \emph{(none)}
  & Painless goiter, hyperthyroid symptoms, normal ESR:
    \textbf{(C) silent thyroiditis}. \\
Nonsense
  & ``\ldots avoid aerodynamic shortcuts that don't generalize\ldots''
  & Painless mass, hyperthyroid symptoms, normal ESR:
    \textbf{(C) silent thyroiditis}. \\
Neutral
  & ``\ldots evidence-based reasoning, avoid common diagnostic errors\ldots''
  & Affects women 30--50, spontaneous resolution:
    \textbf{(C) silent thyroiditis}. \\
DEI
  & ``\ldots account for disparities affecting racial and ethnic
    minorities, LGBTQ+ patients\ldots''
  & Weighs rates in \emph{``Black and Hispanic women''} $\rightarrow$
    \textbf{(D) Hashimoto's thyroiditis}. \\
\bottomrule
\end{tabular}
\end{table}

\begin{table}[t]
\caption{The 47 models by class, with examples and the class-mean injection rate
under the DEI prompt (pooled per class).}
\label{tab:models}
\footnotesize
\setlength{\tabcolsep}{3pt}
\begin{tabular}{@{}p{0.29\columnwidth}c>{\raggedright\arraybackslash}p{0.56\columnwidth}c@{}}
\toprule
\textbf{Class} & \textbf{N} & \textbf{Examples} & \textbf{Inj.} \\
\midrule
Frontier proprietary & 4 & GPT-5, Claude Sonnet 4.6, Grok-4-fast & 49\% \\
Budget proprietary & 11 & GPT-4o-mini, Gemini-flash-lite, Claude Haiku & 34\% \\
Large open ($\geq$65B/MoE) & 6 & Llama-3.3-70B, Qwen2.5-72B, DeepSeek-V3 & 26\% \\
Mid open (10--34B) & 11 & Qwen3-14B, Gemma-3-12B, Phi-4 & 35\% \\
Small open ($<$10B) & 6 & Llama-3-8B, Gemma-2-9B, Mistral-7B & 37\% \\
Medical fine-tune & 9 & OpenBioLLM-8B, MedGemma-4B & 19\% \\
\bottomrule
\end{tabular}
\end{table}

\section{Discussion}
\label{sec:discussion}
\paragraph{Why this matters.} When an input makes the model inject a demographic, it
misrepresents the patient, and that is a problem that can lead to a wrongly matched
recommendations. Demographic injection attaches an unstated attribute to the
patient, so the model reasons about a patient who differs from the one described,
and across all 47 models a small, consistent slice of these changes flips the
selected option to an incorrect answer. A prompt meant to improve equity can thus
reduce the accuracy of the patient description the model works from.

\paragraph{A likely mechanism.} In training text, equity discourse co-occurs with
demographic descriptors, so a model acquires a directional prior that raises the
probability of race- and SES-related continuations when DEI tokens are present.
The rate scales with how explicitly the prompt requests demographics and is gated
by instruction-following, as if the directive were a \emph{content request},
adjacent to sycophancy~\cite{sharma2023sycophancy} and prompt-format
sensitivity~\cite{sclar2024formatting}.

\paragraph{Generalization.} The effect is likely not specific to DEI. Any framing
directive with a strong learned association should elicit that content unbidden
(``consider cost-effectiveness'' injecting insurance framing, ``consider
medico-legal risk'' injecting liability framing), benign until it is
\emph{load-bearing} for the answer. Patient representation is the case to watch in
medicine: other prompts that evoke demographic context can likewise add or shift a
patient's race, age, or social situation, rewriting who the patient is and, as
here, sometimes the recommended answer.

\paragraph{Related work and limitations.} Demographic bias in clinical models is
documented~\cite{omiye2023race,zack2024gpt4,pfohl2024equity}. We instead
study a \emph{prompting practice} as the cause and separate benign from
answer-altering injection at scale, a form of instruction-induced extrinsic
hallucination~\cite{ji2023hallucination}. Limitations: answer-altering rates are
near-floors; gold labels are single-annotator; benchmarks are multiple-choice;
and the mechanism is a hypothesis pending activation-level analysis.

\balance

\end{document}